\documentclass[10pt,twocolumn,letterpaper]{article}

\usepackage{iccv}
\usepackage{times}
\usepackage{epsfig}
\usepackage{graphicx}
\usepackage{amsmath}
\usepackage{amssymb}

\newcommand{\minisection}[1]{\vspace{0.04in} \noindent {\bf #1}\ \ }
\usepackage{listings}  
\usepackage{courier}
\usepackage{amsfonts}
\usepackage{mathtools}  
\DeclareMathOperator*{\argmin}{\arg\!\min}

\usepackage{gensymb}  

\usepackage{graphicx,subfigure}
\usepackage[normalem]{ulem}  
\usepackage{tabu} 

\usepackage{array}  
\newcolumntype{$}{>{\global\let\currentrowstyle\relax}}
\newcolumntype{^}{>{\currentrowstyle}}

\usepackage[pagebackref=true,breaklinks=true,letterpaper=true,colorlinks,bookmarks=false]{hyperref}

\iccvfinalcopy 

\def\iccvPaperID{7} 

\ificcvfinal\fi
\begin{document}

\title{Adversarial Networks for Spatial Context-Aware\\Spectral Image Reconstruction from RGB}

\author{Aitor Alvarez-Gila\\
TECNALIA / CVC - Universitat Aut\`onoma de Barcelona\\
Derio, Spain\\
{\tt\small aitor.alvarez@tecnalia.com}
\and
Joost van de Weijer\\
CVC - Universitat Aut\`onoma de Barcelona\\
Barcelona, Spain\\
{\tt\small joost@cvc.uab.es}
\and
Estibaliz Garrote\\
TECNALIA\\
Derio, Spain\\
{\tt\small estibaliz.garrote@tecnalia.com}
}

\maketitle

\begin{abstract}
   Hyperspectral signal reconstruction aims at recovering the original spectral input that produced a certain trichromatic (RGB) response from a capturing device or observer.
   Given the heavily underconstrained, non-linear nature of the problem, traditional techniques leverage different statistical properties of the spectral signal in order to build informative priors from real world object reflectances for constructing such RGB to spectral signal mapping.
   However, most of them treat each sample independently, and thus do not benefit from the contextual information that the spatial dimensions can provide. 
   We pose hyperspectral natural image reconstruction as an image to image mapping learning problem, and apply a conditional generative adversarial framework to help capture spatial semantics. 
   This is the first time Convolutional Neural Networks -and, particularly, Generative Adversarial Networks- are used to solve this task. Quantitative evaluation shows a Root Mean Squared Error (RMSE) drop of $33.2\%$ and a Relative RMSE drop of $54.0\%$ on the ICVL natural hyperspectral image dataset.
\end{abstract}


\section{Introduction}
\label{sec:intro}

Hyperspectral (HS) imaging has gained relevance over the last couple of years in the applied vision community. 
Remote sensing, UAV-based imaging, precision agriculture or autonomous driving are only some of the fields that are already benefiting from the use of imaging devices that provide a response that spans the spectral dimension with narrow-band channels to produce an image with higher spectral resolution than the standard RGB trichromatic one.

While the evolution of HS imaging devices has undergone major breakthroughs, it is also true that there is still a trade-off inherent to the fact that we are ultimately capturing three dimensional information with a two dimensional sensor, which limits the quality or resolution of the acquired signal in either of those dimensions: spatial, spectral or temporal. 
On top of that, the cost of such devices is orders of magnitude above that of conventional RGB cameras.

In this context, HS signal reconstruction from broadband or limited acquisition channels (typically, from RGB sensors) arises as a natural computational alternative, either to compete against native HS systems or to be included as part of their signal post-processing backends. 
The spectral reconstruction problem is a severely underconstrained, highly non-linear one, and the algorithms trying to solve this mapping should exploit the low dimensionality of the natural HS images~\cite{chakrabarti2011statistics} and learn informative priors of diverse forms from real world object reflectances, to be leveraged in the reconstruction phase.
Note, however, that most of the existing solutions handle each pixel individually.
By doing so, they are not taking advantage of the latent contextual information available in the spatially local neighborhood~\cite{chakrabarti2011statistics}.

Generative adversarial Networks (GAN) are a class of neural networks which have shown to be able to successfully generate samples from the complex manifold of real images.
In this work, we use this class of algorithms to learn a generative model of the joint spectro-spatial distribution of the data manifold of natural HS images and use it to optimally exploit spatial context information.
To our knowledge, this is the first time Convolutional Neural Networks (CNN) are used in the task of spectral reconstruction of natural images.
We quantitatively evaluate our approach on the largest HS natural image dataset available to date, i.e. ICVL, by comparing against~\cite{arad_sparse_2016}, and show error drops of $33.2\%$ (RMSE) and $54.0\%$ (relative RMSE) over their state of the art results.

\subsection{Related work}
\label{sec:related}

A number of works are relevant to the proposed approach.
This task was first addressed by isolating its spatial component and focusing on the reconstruction of homogeneous, well-established reflectances of real world surfaces such as Munsell chips, either from multispectral, RGB components~\cite{heikkinen_evaluation_2008} or from the tristimulus values~\cite{ayala_use_2006,agahian_reconstruction_2008}.

Initial attempts on the spectral reconstruction of natural images from full size RGB input required additional constrains or multiple input forms to help in their task:~\cite{kawakami_high-resolution_2011} and~\cite{cao_high_2011} use the aid of a low resolution HS measurement in addition to the RGB input,~\cite{lopez-alvarez_using_2008} restricts to the skylight samples domain, and~\cite{park_multispectral_2007,parmar_spatio-spectral_2008,goel_hypercam:_2015}, among others, rely on the aid of computational photography-like multiplexed narrow band lighting. The latter does, however, use spatial information for learning, as does~\cite{chakrabarti2011statistics}, which focuses on the statistics for this class of images and defines a representation basis and computation method for the associated coefficients, but does not tackle reconstruction.

Solutions relying on a single RGB image input at test time are scarce, and almost none of them leverage the spatial context:~\cite{nguyen_training-based_2014} uses a Radial Basis Function network and produces an estimate of scene reflectance and global illuminant, but assumes a known camera color matching function, and directly depends on the performance of a white balancing stage as part of the workflow.
\cite{zhao_image-based_2007} presents the \emph{matrix R method} for spectral reflectance reconstruction, which additionally requires a calibration target to build a camera model.
\cite{arad_sparse_2016} learns a sparse dictionary of HS signatures as bases for the reconstruction.
By treating each pixel independently, the ability to use the surround information is lost e.g. for producing distinct spectral outputs for metameric RGB pairs dependent on the context.

Remarkably,~\cite{RoblesKelly2015SingleIS} exploits spatial material properties of the imaged objects by extracting not only spectral, but also convolutional features resulting from the application of the filter banks from~\cite{varma_classifying_2002}, and adopting a constrained sparse coding-based reconstruction approach. In parallel to our development, we found a similar approach~\cite{galliani_learned_2017} which makes use of a CNN-based encoder-decoder to address this task.

Finally, there exists a certain relation between the HS reconstruction and the image colorization~\cite{Cheng_2015_ICCV} tasks, which has been previously addressed in a similar fashion~\cite{zhang_colorful_2016,Isola_2017_CVPR}, but under different evaluation requirements.
We can think of the former being a generalization of the latter for an arbitrary number of input/output channels. 

None of these methods would have been possible without the existence of publicly available HS natural image datasets. Until recently, the amount of images per set was the limiting factor for the development of HS reconstruction algorithms that learn on the basis of images or image patches
\cite{foster_frequency_2006,yasuma_generalized_2010,chakrabarti2011statistics,nguyen_training-based_2014,eckhard_outdoor_2015,foster_time-lapse_2016}.~\cite{arad_sparse_2016} changed this releasing a set of $201$ high resolution images that we show is enough for the successful training of deep neural networks.   

\begin{figure*}
	\begin{center}
		\includegraphics[width=1.0\linewidth]{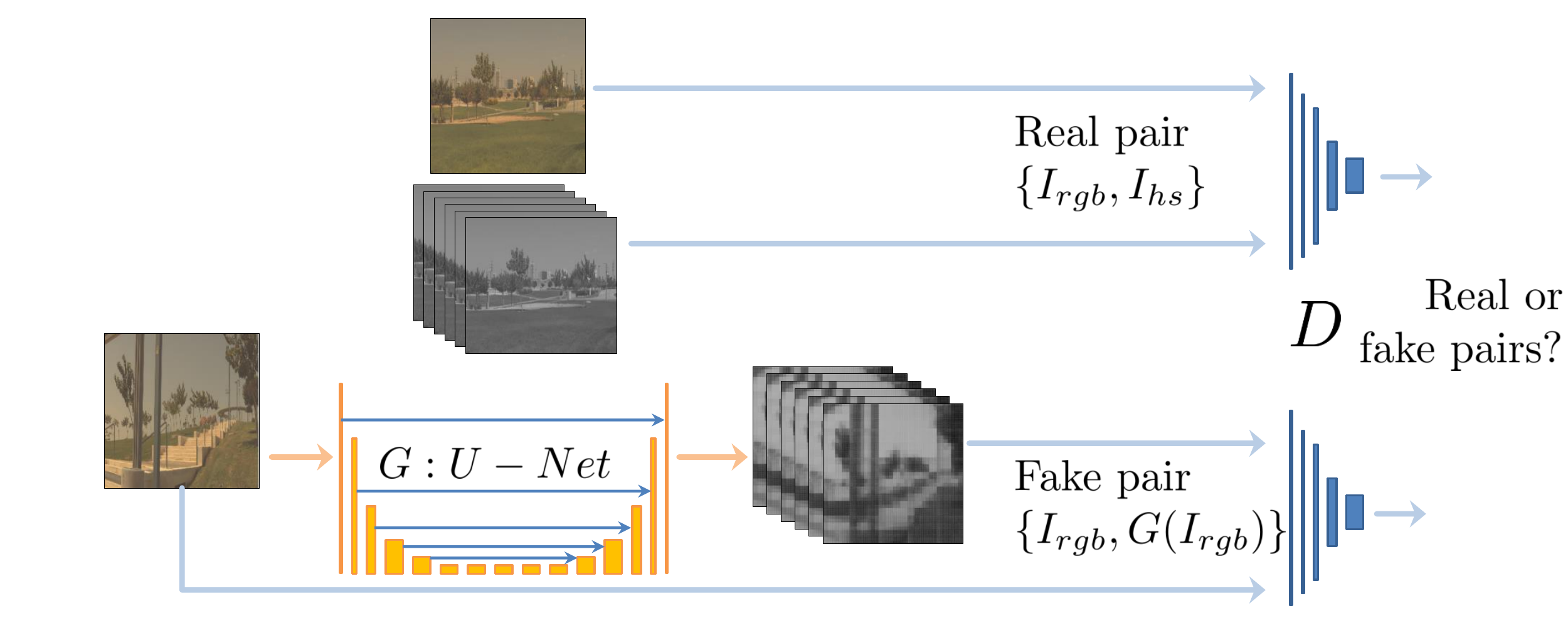}
	\end{center}
	\caption{Adversarial spatial context-aware spectral image reconstruction model}
	\label{fig:overview}
\end{figure*}

\section{Adversarial spectral image reconstruction from RGB}
\label{sec:adv_rgb2hs}

This section describes the core functioning of our method, along with some of the mathematical developments that derived into the proposed models.
\subsection{Adversarial learning}

\minisection{Generative Adversarial Networks (GANs)} 
GAN-s \cite{goodfellow_generative_2014} are generative statistical models that learn to produce realistic samples $y$ that lay in the data manifold by relying on a setup consisting on two competing agents: the generator $G$ takes  noise $z$ as input as a source of randomness, and creates \emph{fake} data samples $G(z)$.
It is trained to make the generated samples as realistic as possible.
On the other end, the aim of the discriminator, $D$, which randomly takes as input both samples from the training data set and those generated by $G$, is to learn to tell if the received input samples are real or fake.
Typically, both $G$ and $D$ are neural nets, and they are trained iteratively to progressively become better in their respective tasks.
The objective function associated to such a setting is:
\begin{align} \label{eq:l_gan}
\mathcal{L}_{GAN}(G,D) &= \mathop{\mathbb{E}_{y\sim p_{data}(y)}} [\log D(y)] + \nonumber \\
					   &+ \mathop{\mathbb{E}_{z\sim p_{noise}(z)}} [\log(1-D(G(z)))] 
\end{align}
where $G$ tries to minimize this loss and $D$ attempts to maximize it, yielding  the objective function:
\begin{equation} \label{eq:minmax_lgan}
G^* = \argmin_{G} \max_{D} \mathcal{L}_{GAN}
\end{equation}
This adversarial framework has successfully been applied to the unsupervised generation of data of different modalities, including natural images~\cite{denton_deep_2015}, and empirical architecture guidelines for $G$ and $D$ have been derived~\cite{radford_unsupervised_2016} for such cases, along with common tricks to stabilize the training process~\cite{salimans_improved_2016}.

\minisection{Conditional Generative Adversarial Networks (cGANs)} 
cGANs~\cite{mirza2014conditional} extend this framework by feeding both $G$ and $D$ with additional information $x$ to be used to condition on the output of the generator. 
Such conditioning input could adopt different modalities, and range from simple categorical labels~\cite{mirza2014conditional} to more sophisticated content, such as text~\cite{reed_generative_2016}  or images~\cite{Li2016}, either alone or as a combination of multiple input  modalities~\cite{reed_learning_2016,zhu_generative_2016}. 
This has been proved useful for a number of tasks and output types~\cite{wang_generative_2016, mathieu2015deep}.
Eq.~\ref{eq:l_cgan} shows the updated loss function for conditional GANs. In this case, $G$ attempts to generate images that look realistic given the additional provided input $x$ (be it the class of $y$, a descriptive text, or an additional image), and $D$ tries to determine whether the given $(x,y)$ pair makes sense or not as a mapping.   
\begin{align} \label{eq:l_cgan}
&\mathcal{L}_{cGAN}(G,D) = \nonumber \\
&= \mathop{\mathbb{E}_{x,y\sim p_{data}(x,y)}} [\log D(x,y)] + \nonumber \\ 							&+\mathop{\mathbb{E}_{x\sim p_{data}(x),z\sim p_{noise}(z)}} [\log(1-D(x, G(x,z)))]
\end{align}
As a result, cGANs open the door to using generative statistical modeling for our HS reconstruction problem by conditioning the generation of an HS outcome on a given input RGB image.

\minisection{Adversarial image to image mapping}
Many modern computer vision tasks can better be regarded under the common reference framework of image to image mapping learning, in which a generator model $G$ is learned that translates an input image $x$ into the most probable representation $y$ of such image in the output domain.
This is the case \eg for semantic segmentation~\cite{shelhamer_fully_2016}, instance segmentation~\cite{Dai_2016_CVPR}, or depth and surface normal estimation from single image~\cite{bansal_pixelnet:_2017}, among others. 
Most of these tasks have been recently addressed making use of Convolutional Neural Networks that yield deterministic results as generators, and which are specifically tailored, in terms of architecture design, objective function or other specific training details, for their respective tasks. 

There are, in addition, some tasks for which this mapping is not unique, and one same input image could have multiple equally correct representations in the output domain.
Realistic image rendering from semantically labeled images (inverse of the semantic segmentation problem) or from hand-drawn sketches, or image colorization~\cite{Cheng_2015_ICCV}, are just a few examples of this. 
The choice of the objective functions to use in each of these cases is a particularly challenging design aspect; applying an otherwise useful $\ell_2$ loss to $x,y$ image pairs is known to be problematic and yield blurry results~\cite{larsen_autoencoding_2016}, as the generator tends to average over the space of valid image representations.  

For all of the above,~\cite{Isola_2017_CVPR} proposes a common image to image mapping learning framework based on the cGAN adversarial setting, which, provided that one can feed it with co-registered image pairs of input and output domains, is able to learn the most suitable loss function for each of the tackled tasks in a data-driven approach. This is done implicitly using the adversarial objective from eq.~\ref{eq:l_cgan}, enforced by the discriminator trying to identify the fake images and, this way, encouraging the generator to become better at trying to deceive it. 

By doing this,~\cite{Isola_2017_CVPR} manages to get rid of the blur inherent to $\ell_2$ distance-based models and produce sharp results.
Nevertheless, it has been previously shown~\cite{Pathak_2016_CVPR, Shrivastava_2017_CVPR} that combining one of the traditional loss functions with the adversarial objective $\mathcal{L}_{cGAN}$ can help produce more spatially consistent results and make the generator less prone to artifacts inherent to the adversarial scheme.
They thus place an additional $\ell_1$ term (eq.\ref{eq:ll1}) on the generator, which is known to yield less blur:
\begin{equation} \label{eq:ll1}
\mathcal{L}_{\ell_1}(G) = \mathop{\mathbb{E}_{x,y\sim p_{data}(x,y),z\sim p_{noise}(z)}} [\|y-G(x,z)\|_1]
\end{equation}
and produce the following combined objective function:
\begin{equation} \label{eq:minmax_lgan_lone}
G^* = \argmin_{G} \max_{D} \mathcal{L}_{cGAN}(G,D) + \lambda\mathcal{L}_{\ell_1}(G)
\end{equation}
where $\lambda$ is a weighting factor for the $\ell_1$ term, which is set to 100 in~\cite{Isola_2017_CVPR}. 
In essence, $\mathcal{L}_{cGAN}(G,D)$ would be in charge of producing sharp, realistic looking results, while $\ell_1$ takes care of the global image structure.

Interestingly, the stochastic output pursued by the noise input to cGAN-like models does not manifest itself under this design (see details in section \ref{sec:architecture}), and the resulting mapping is a fundamentally deterministic one.
A probable interpretation is $G$ learning to ignore the effect of the noise. 
As a result, ~\cite{Isola_2017_CVPR} gets rid of the noise input and leaves test-time dropout as unique source of randomness.

\minisection{Adversarial spectral reconstruction networks} The forward correspondence learning between the RGB and hyperspectral signals is a heavily under-constrained one, which could benefit from an approach that aims at exploiting the underlying priors present in both the spectral and spatial dimensions and learn a model that specifically produces realistic outcomes as a target. 
It not only requires mapping a 3-dimensional image to a  much higher dimensional one (typically 31 spectral channels and the two spatial dimensions), but such mapping can be context-dependent as well, as is in the case of metameric colors.
The inverse mapping, however, i.e. the rendition of RGB images from their spectral counterparts, is well defined, and deterministic under the only assumption of the color matching functions defining the observer, or the spectral sensitivity functions that characterize specific sensors.
This makes it immediate to generate perfectly aligned (RGB, hyperspectral) image pairs (see section \ref{sec:experimental_evaluation}) to be used under the described solution.

Hyperspectral image reconstruction from RGB can then be posed as one of the aforementioned image to image mapping learning problems and thus be solved under the conditional adversarial network-based image to image translation framework proposed by~\cite{Isola_2017_CVPR}.

The resulting adversarial and combined objectives would then become:
\begin{multline} \label{eq:l_hscgan}
\mathcal{L}_{adv} = \mathop{\mathbb{E}_{I_{rgb},I_{hs}\sim p_{data}(I_{rgb},I_{hs})}} [\log D(I_{rgb}, I_{hs})] + \\
+ \mathop{\mathbb{E}_{I_{rgb}\sim p_{data}(I_{rgb})}} [\log(1-D(I_{rgb},G(I_{rgb})))]
\end{multline}
\begin{multline} \label{eq:ltot}
\mathcal{L}_{rgb2hs}(G,D) = \mathcal{L}_{adv} + \lambda\mathcal{L}_{\ell_1} = \mathcal{L}_{adv} + \\
+ \lambda\mathop{\mathbb{E}_{I_{rgb},I_{hs}\sim p_{data}(I_{rgb},I_{hs})}} [\|I_{hs}-G(I_{rgb})\|_1]
\end{multline}
where $I_{hs}$ represents the original hyperspectral image, $I_{rgb}$ is the corresponding input image in the RGB domain and $\lambda$ is scalar weight used to balance both loss terms (and is set to 100 in all our experiments, unless otherwise stated). Note that we have explicitly removed any reference to the input noise, and the RGB image remains as the only input to $G$. 

Figure \ref{fig:overview} shows an overview of the whole adversarial spatial context-aware spectral image reconstruction process. We depart from a database of perfectly aligned RGB and hyperspectral image pairs, which are extracted one pair at a time. In a first iteration, a first pair of real images of size $H\times W$ is taken: $\{I_{RGB}, I_{HS}\}$. The generator $G$ takes $I_{RGB}$ as input, and yields the corresponding hyperspectral reconstruction of size $H\times W$, $\hat{I}_{HS}$. The discriminator $D$ is now fed with two pairs of images, $\{I_{RGB}, I_{HS}\}$ and $\{I_{RGB}, \hat{I}_{HS}\}$ and uses the associated labels indicating if they are real or fake $\{1,0\}$ to compute the adversarial loss and update its gradients. $G$'s weights are also updated, and both $D$ and $G$ continue to become better at their respective tasks iteratively.

\begin{figure*}[!htbp]
	\centering     
	\subfigure{\label{fig:srgb_samples1}\includegraphics[width=0.13\textwidth]{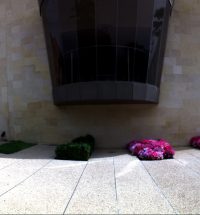}}
	\subfigure{\label{fig:srgb_samples2}\includegraphics[width=0.13\textwidth]{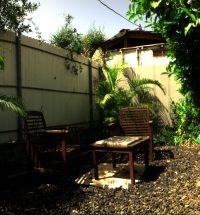}}
	\subfigure{\label{fig:srgb_samples3}\includegraphics[width=0.13\textwidth]{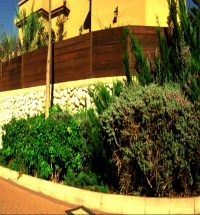}}
	\subfigure{\label{fig:srgb_samples4}\includegraphics[width=0.13\textwidth]{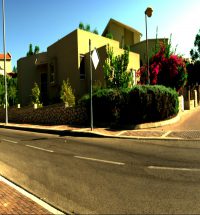}}
	\subfigure{\label{fig:srgb_samples5}\includegraphics[width=0.13\textwidth]{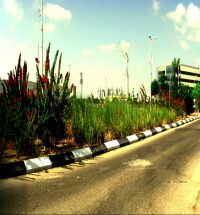}}
	\subfigure{\label{fig:srgb_samples6}\includegraphics[width=0.13\textwidth]{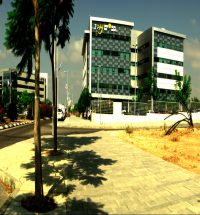}}\\
	\subfigure{\label{fig:srgb_samples7}\includegraphics[width=0.13\textwidth]{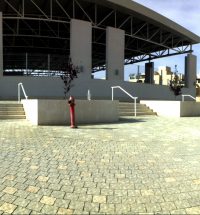}}
	\subfigure{\label{fig:srgb_samples8}\includegraphics[width=0.13\textwidth]{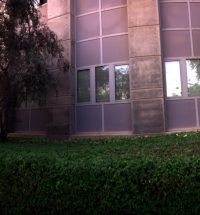}}
	\subfigure{\label{fig:srgb_samples9}\includegraphics[width=0.13\textwidth]{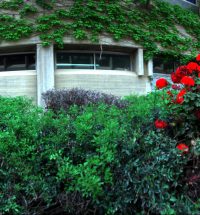}}
	\subfigure{\label{fig:srgb_samples10}\includegraphics[width=0.13\textwidth]{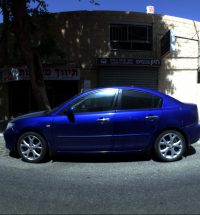}}
	\subfigure{\label{fig:srgb_samples11}\includegraphics[width=0.13\textwidth]{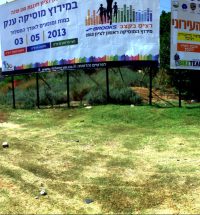}}
	\subfigure{\label{fig:srgb_samples12}\includegraphics[width=0.13\textwidth]{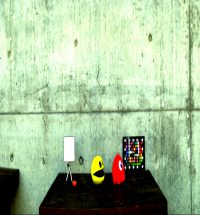}}
	\caption{Random RGB samples from the ICVL dataset~\cite{arad_sparse_2016}.}
	\label{fig:srgb_samples}
\end{figure*}

\subsection{Architecture design and training}
\label{sec:architecture}

As for the specific implementation of the models, since $G$ needs to yield full-size detailed images, a \emph{U-Net}-like architecture~\cite{ronneberger_u-net:_2015} is used.
Regular autoencoder networks~\cite{kingma_auto-encoding_2013} exhibit a progressively reduced representation size until a bottleneck layer and there is no way for the last layers of accessing the original data, which negatively affects the results when we aim at detailed outcomes. 
Unlike these, the \emph{U-Net} incorporates skip connections between layers of equal representation size, and concatenates local activations from the upscaling phase with those coming from the downscaling stages, which has shown to achieve superior performance on tasks were the details are relevant.
It was first proposed with semantic segmentation tasks in mind, but original spectral signal reconstruction falls within the kind of tasks that can clearly benefit from accessing the original input  levels at each sample (i.e. pixel).

The discriminator $D$, defined as \emph{PatchGAN}, is simpler in terms of convolutional layer count, and is focused solely on modeling high-frequency structure. Each of the $M\times M$ output neurons is restricted to see only a limited $N\times N$ receptive field from the input image, which can be significantly smaller than the input image size.
Consequently, only the adversarial loss term is placed over $D$ (eq \ref{eq:minmax_lgan_lone}).

The use of this design solution for $D$ is consistent with our initial hypothesis that local spatial context can help better reconstruct the spectral signal. 
Specifically, we hypothesize that the proposed approach could help disentangling the illuminant and object body reflectance components of a pixel's trichromatic response, as defined by the dichromatic reflection model \cite{shafer_using_1985}.
The design of $D$, with its attached $\ell_1$ objective, helps capture the high frequencies that characterize the textures in the image.
These are, together with the body color component, one of the main features characteristic of the different materials which, ultimately, produce distinct spectral responses. Therefore, convolutionally integrating the trichromatic response of adjacent pixels should yield a better estimate of the central spectral response. 
To this respect, the \emph{PatchGAN} design isolates $D$'s response associated to pixels separated by more than one input patch.
For small enough patch sizes, this effectively implies that the discriminator is learning a loss function tailored for texture or material recognition, making sure that the reconstructed spectra falling within the patch are not only plausible in the spectral domain, but also spatially consistent in the close proximities. 

The illuminant-specific component of ~\cite{shafer_using_1985}, on the other hand, is typically largely constant or slowly varying across big portions of the image (especially in terms of chromaticity and conversely, spectral shape), and the $\ell_1$ norm does a good job taking care of its global image-wide consistency, along with that of the low-mid frequency spatial structures. 

\minisection{Avoiding Batch Normalization}
Given the intrinsically exact nature of our task (some of the described design choices help leverage spatial structure consistency for our task, but we ultimately want the reconstructed spectra to be accurate), we choose to remove all the Batch Normalization~\cite{ioffe_batch_2015} layers present in the generator architectures proposed in~\cite{Isola_2017_CVPR}.
While this technique has shown to be useful to help accelerate and regularize the training process for a wide variety of tasks by reducing the internal covariate shift, the fact that it makes the signal lose track of its original value, along with the deterministic nature of the desired output, makes it non-advisable for reconstruction tasks. 
We experimentally found that including Batch Normalization produced inferior results.

\subsection{Implementation details}
\label{sec:implementation}
We now provide some details on the configurations used for our implementation. We use Keras with Theano backend and take the implementation of~\cite{Isola_2017_CVPR} made by~\cite{costa_towards_2017} as starting point, modifying it for our purposes.
We use Adam optimizer~\cite{kingma_adam:_2015} for both $G$ and $D$, with a learning rate of $2\cdot10^{-4}$ and $\beta_1=0.5$. We use a minibatch size of 1 in order to benefit from the regularization provided by the gradient estimation noise~\cite{keskar_large-batch_2017}, and following common practice~\cite{Isola_2017_CVPR}. The training is performed iteratively and alternates between the two models: at each step, the discriminator is first trained for $50$ iterations and then the generator gets trained for $25$ more minibatches.

We crop the original $1392\times 1300$ images during the training phase by extracting one random crop of size $256\times 256$ (the $H,W$ values from section \ref{sec:adv_rgb2hs}) per image and epoch.
The models are fed with these crops during training, while, for the testing phase, each full size RGB image is divided in tiles of $256\times256$ with no overlap, which effectively yields image sizes of $1280\times1280$ pixels.
Each tile gets processed by the generator independently and we reconstruct the full image back before evaluating it.

The generator $G$ accepts input images of size $256\times 256$. Its encoding stage is composed by eight successive $3\times3$ convolutions with stride 2 and a leaky ReLU after each of them, thus yielding a $1\times1$ activation in the most narrow point of the main branch. The initial number of filters is $64$, which gets doubled at each convolutional layer up to $512$, keeping it constant after that. On the decoding part, eight transposed convolution blocks successively double the activation size up until the original $256\times256$ size, while progressively reducing the number of filters in a symmetric way with respect to the encoding stage. Each block comprises the transposed convolution itself, followed by a train-time-only Dropout layer (with a drop rate of $10\%$) and a leaky ReLU activation. After each Dropout, the correspondent activations from the encoding stage are concatenated, thus producing eight skip connections between levels of equivalent activation size. Finally, two $1\times1$ convolutions are added at the end before the output \emph{tanh} activation, with a leaky ReLU in between, in order to get the direct input images adequately combined with the upstream features.

The discriminator $D$ is a simple single-branch net composed of four $3\times3$ convolutional layers with stride 2, each of them followed by a leaky ReLU, with filter numbers doubling at each step. A fifth $3\times3$ convolution with a sigmoid yields the output $8\times8$ prediction.

\section{Experimental evaluation}
\label{sec:experimental_evaluation}

This section contains an overview of the experiments performed to quantitatively assess our algorithm's performance as compared to previous methods.

\subsection{Dataset}
Given the amount of images, diversity and resolution, we evaluate our approach on the dataset presented in~\cite{arad_sparse_2016}. At the time of writing, it comprised $201$ hyperspectral images (see Figure~\ref{fig:srgb_samples} for RGB renditions of a few random samples) of $1392\times1300$ spatial resolution and $519$ spectral bands in the $400nm-1000nm$ range, with a spectral resolution of 1.25nm. 
As for the acquisition, a \emph{Specim PS Kappa DX4} hyperspectral camera was used, together with a rotary stage for spatial scanning.
This aspect is noticeable in some of the samples, in which common objects such as cars exhibit aspect ratios that do not match those we find in real life. There is also a spectrally downsampled version of $31$ bands in the $400nm-700nm$ range. Following practice from~\cite{arad_sparse_2016}, we use the latter for our reconstruction experiments. There is no illuminant information available for each of the images, which would allow for object reflectance recovery; therefore, our task consists on the estimation of the radiance correlate represented by the captured hyperspectral images. 

\subsection{Preparation}
\label{ssec:preparation}

In order to get the aligned image pairs dataset required by our method, and given the deterministic correspondence between spectral and RGB samples once the observer (or sensor sensitivity functions) and the output color space are specified, we render wide band trichromatic RGB versions of the spectral images in the sRGB color space as follows: we first obtain the CIE $XYZ$ tristimulus values for each spectral image pixel location $x$, making use of the color matching functions corresponding to the CIE 1964 $10\degree$ standard observer:
\begin{equation} \label{eq:trix}
\mathbf{X}(x)=K(x)\sum_{\lambda=400nm}^{700nm}S(\lambda, x)\mathbf{\bar{x}(\lambda)}\Delta\lambda
\end{equation}
where $S(\lambda, x)$ is the relative spectral power distribution of pixel $x$, $\mathbf{X}=\{X,Y,Z\}$, $\mathbf{\bar{x}(\lambda)}=\{\bar{x}(\lambda), \bar{y}(\lambda), \bar{z}(\lambda)\}$ are the color matching functions, $\Delta\lambda=10nm$ and $K(x)$ is the normalization factor, defined, for illuminant $L(\lambda, x)$, as:
\begin{equation} \label{eq:trik}
K(x)=\frac{100}{\sum_{\lambda=400nm}^{700nm}L(\lambda, x)\bar{y}(\lambda)\Delta\lambda}
\end{equation}
Note that, before going through this computation, the original spectral power distribution captured by the camera for each image $S'(\lambda, x)$ is preprocessed with min value subtraction and max value scaling.
The final step is producing the sRGB renders. We do so by applying the associated $3\times3$ transformation matrix and unlinearizing (i.e. \emph{gamma-correcting}) the result with a $1/2.4$ power law gamma with a linear segment in low luminance values.

While not suffering from the same lack of an adequate performance evaluation method that affects typical generative modeling tasks~\cite{theis_note_2016}, spectral signal reconstruction algorithms assessment is an active research field that lacks consensus on what is the most adequate metric to measure spectral match of signals~\cite{imai_comparative_2002}.
When the signals comprise the visual spectrum, the task can be tackled from a variety of perspectives, ranging from the pure signal processing point of view of spectral curve difference metrics, to a full spectrum of metric families that place different levels of perceptual load on their computation: metameric indexes, CIE $\Delta E$ color difference equations, or weighted spectral metrics.

If we widen the scope onto full reference image difference metrics, little work has been done on the spectral extension of these families~\cite{moan_image-difference_2014}. We here focus on four of the most widely used metrics, namely RMSE (Root Mean Squared Error, computed across the spectral dimension for each pixel and then averaging for whatever number of pixels present in the image or the dataset), RMSERel (i.e. RMSE relative to the value of the real signal), GFC (Goodness of Fit Coefficient \cite{romero1997linear}) and~$\Delta E_{00}$ (CIEDE2000) perceptual color difference formula~\cite{cie_cie_2001} computed over the reconstructed tristimulus values.
\subsection{Experiments and discussion}
\begin{table*}[h]
	\begin{center}
       	\begin{tabu} to \linewidth {$c|^c^c^c^c}
			\hline
			Method & RMSE & RMSERel & GFC & $\Delta E_{00}$\\
			\hline
            Galliani \etal\cite{galliani_learned_2017} (reported) & 1.980 & 0.0587 & $-$ & $-$\\
			\hline
			Arad \etal\cite{arad_sparse_2016} (reported) & 2.633 & 0.0756 & - & - \\
            Arad \etal\cite{arad_sparse_2016} (optimized parameters) & $2.184\pm 0.064$ & $0.0872\pm 0.004$ & $-$ & $-$\\
            \hline
            \textbf{Ours (weighted avg.)}  & $\mathbf{1.457\pm 0.040}$ & $\mathbf{0.0401\pm 0.0024}$ & $\mathbf{0.99921\pm 0.00012}$ & $\mathbf{2.044\pm 0.341}$\\
			Ours (fold 0) & $1.452\pm 0.101$ & $0.0383\pm 0.0024$ & $0.99906\pm 0.00001$ & $1.861\pm 0.324$\\
			Ours (fold 1) & $1.463\pm 0.022$ & $0.0420\pm 0.0024$ & $0.99936\pm 0.00023$ & $2.228\pm 0.358$\\
			\hline
		\end{tabu}
	\end{center}
	\caption{Summary results of the conducted experiments over ICVL dataset. Black pixels contained in the original hyperspectral images (derived from the variable image width) are not taken into account for evaluation purposes in any of the experiments, and folds are weighted accordingly. RMSE values are in the $[0-255]$ range. Two train-test cycles were run and the results averaged.}
	\label{tab:results}
\end{table*}
\begin{figure*}[h]
	\begin{center}
        \includegraphics[width=0.495\linewidth]{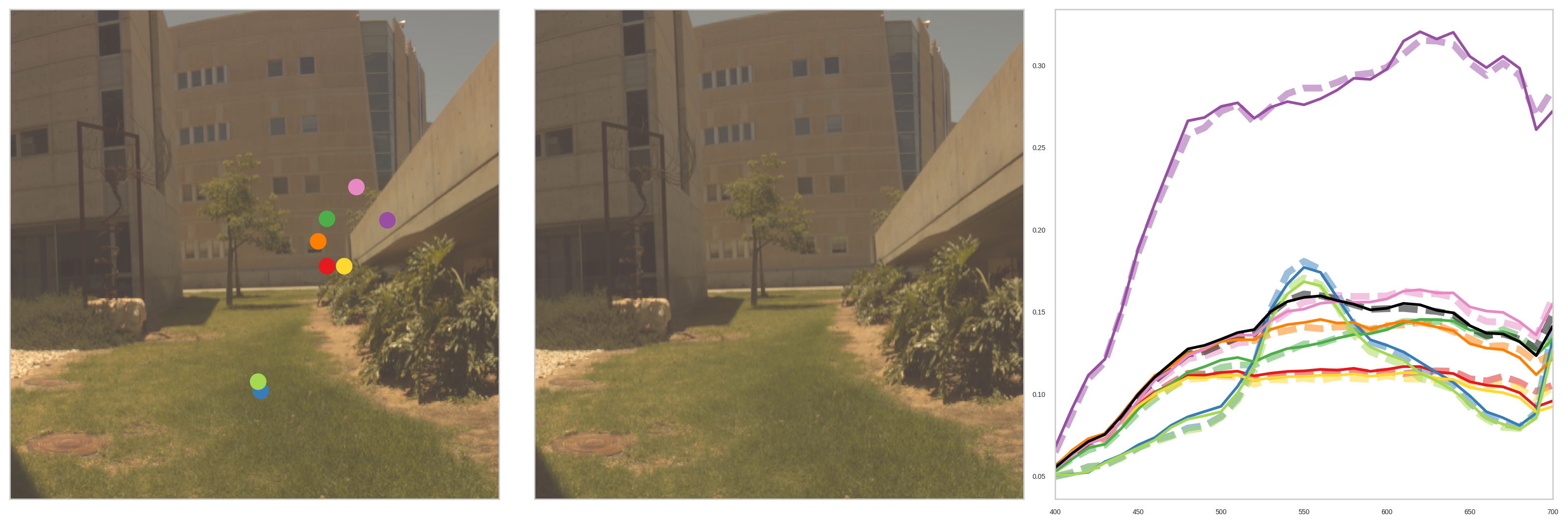}
		\includegraphics[width=0.495\linewidth]{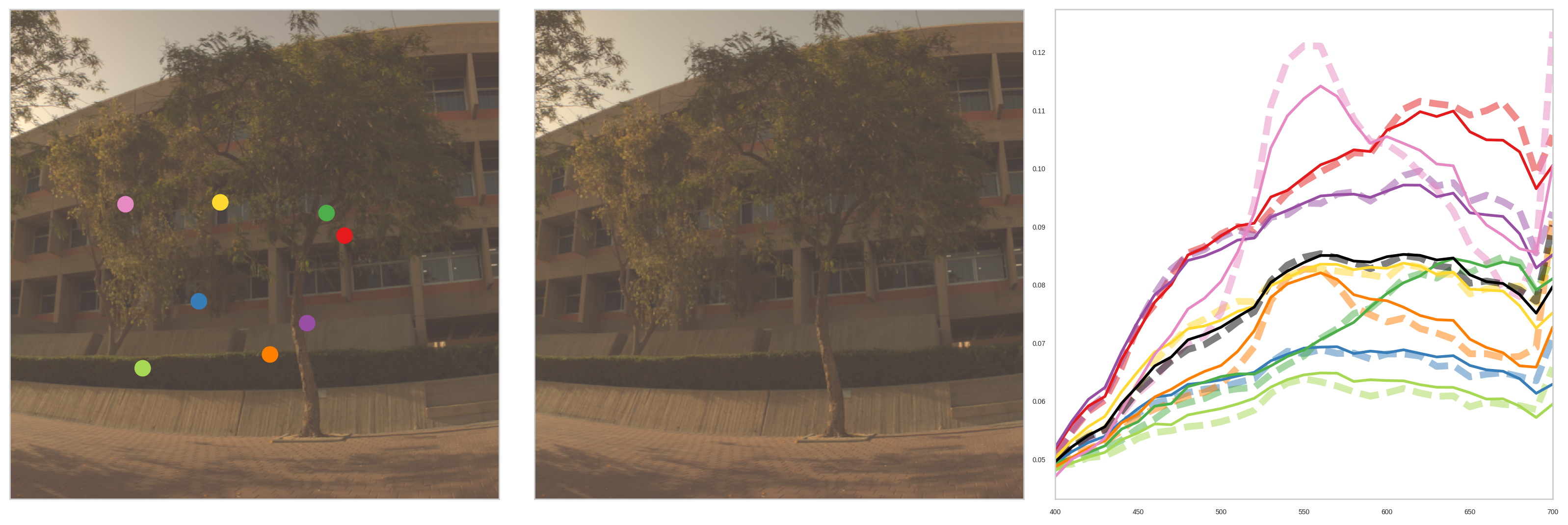}
		\includegraphics[width=0.495\linewidth]{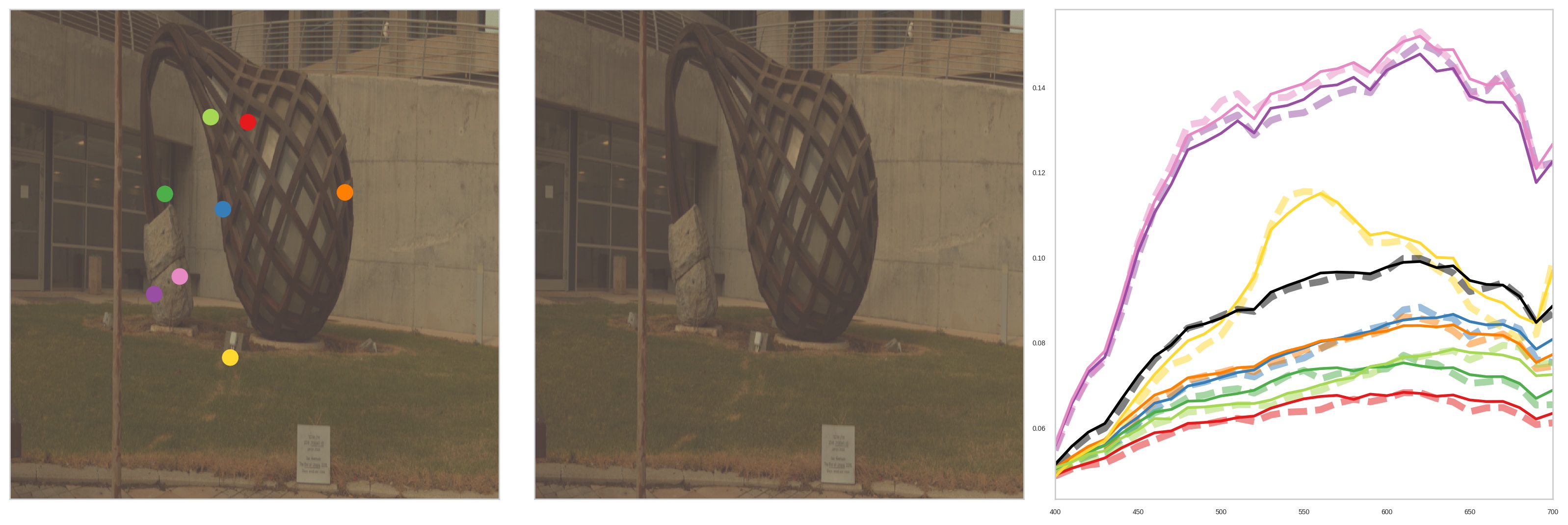}
		\includegraphics[width=0.495\linewidth]{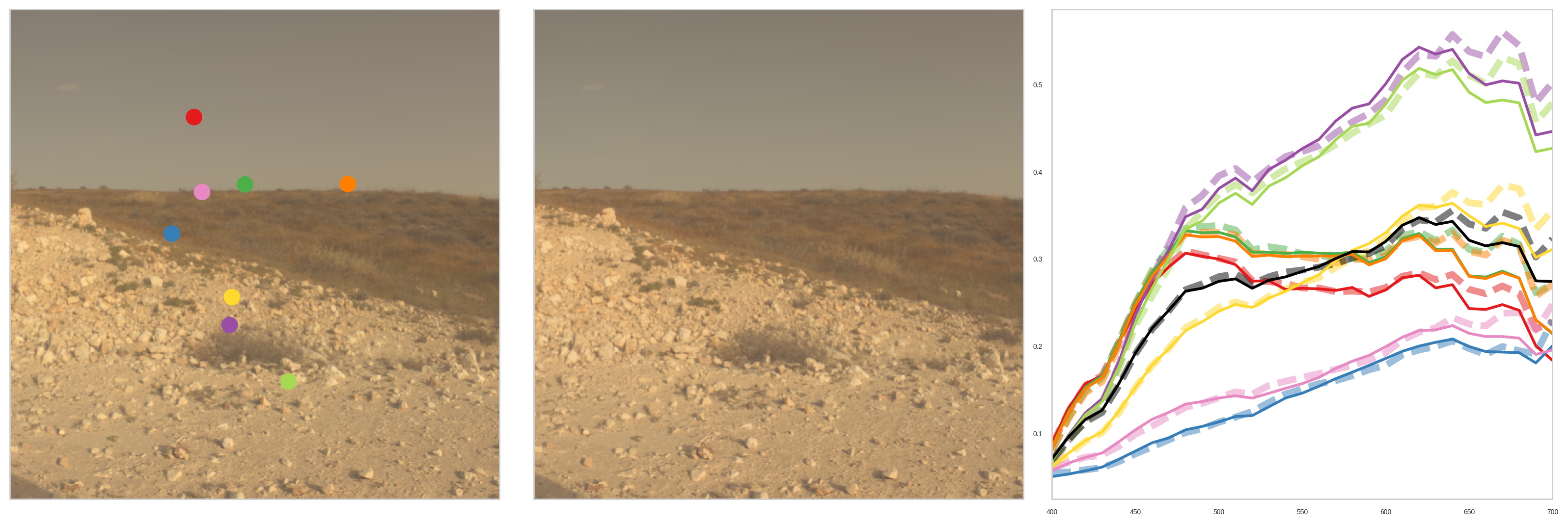}
		\includegraphics[width=0.495\linewidth]{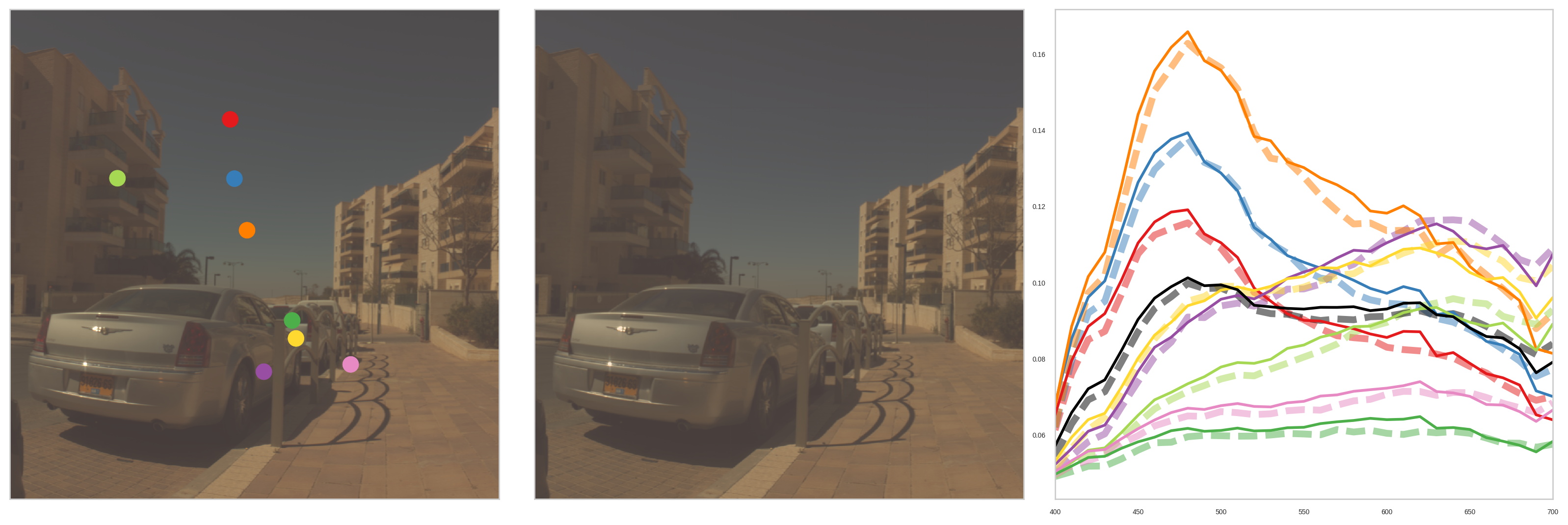}
		\includegraphics[width=0.495\linewidth]{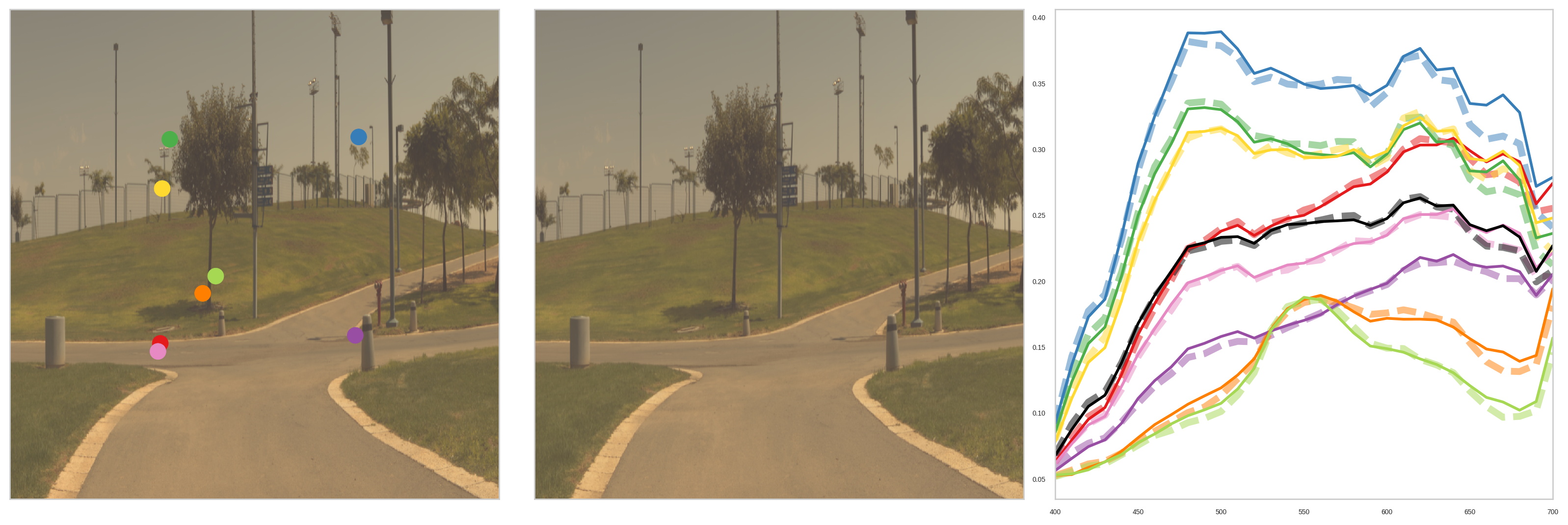}
	\end{center}
	\caption{Sample results for our method. For each triplet, left, center: sRGB rendition of original and reconstructed hyperspectral signals, respectively. Right: Original (dashed) and reconstructed (solid) spectra of eight random pixels identified by the colored dots.}
	\label{fig:results}
\end{figure*}

We first compare our method with~\cite{arad_sparse_2016}. In their general experiment over the whole set, which was back then composed of $100$ images, they perform a leave-one-out procedure, and learn from pixels sampled along the whole set except for the unique image being tested at a time. Their reported results for this setting are shown in Table~\ref{tab:results} as \emph{Arad \etal (reported)}, together with those reported in\cite{galliani_learned_2017} on their own evaluation setting. We choose to instead split the full dataset in two equal partitions of $100$ images each\footnote{Train-test splits available at \url{https://aitorshuffle.github.io/publication/2017-10-10-alvarez-gila_adversarial_2017} and in the supplementary materials section}, training on one and reporting on the other, running two full train-test cycles and averaging the results across folds and runs. Table~\ref{tab:results} compares the obtained values for the aforementioned metrics over each of the testing sets, showing an average per-pixel error drop of $33.2\%$ in terms of RMSE and $54.0\%$ in terms of RMSERel with respect to~\cite{arad_sparse_2016} evaluated over the same splits and with their hyperparameters (i.e. dictionary size, number of samples per image, iterations and sparsity target) optimized over the test sets. While~\cite{arad_sparse_2016} does not provide any further evaluation metric, note that our average GFC values are all above the GFC threshold which~\cite{romero1997linear} considers a \emph{very good reconstruction}, and one which implies missing only $0.2\%$ of the signal energy in the process. Also, the average per-pixel color difference (which does not account for spatial perceptual effects) is constrained around as low as $2\Delta E_{00}$ units.

Figure~\ref{fig:results} shows the sRGB rendition of original and reconstructed hyperspectral images for some randomly chosen test image samples. In addition, for each image, we show the original and estimated spectra for eight randomly selected pixels from the image.

\subsubsection{Does the spatial information actually help?}
\begin{figure*}[!h]  
	\begin{center}
		\includegraphics[width=0.495\linewidth]{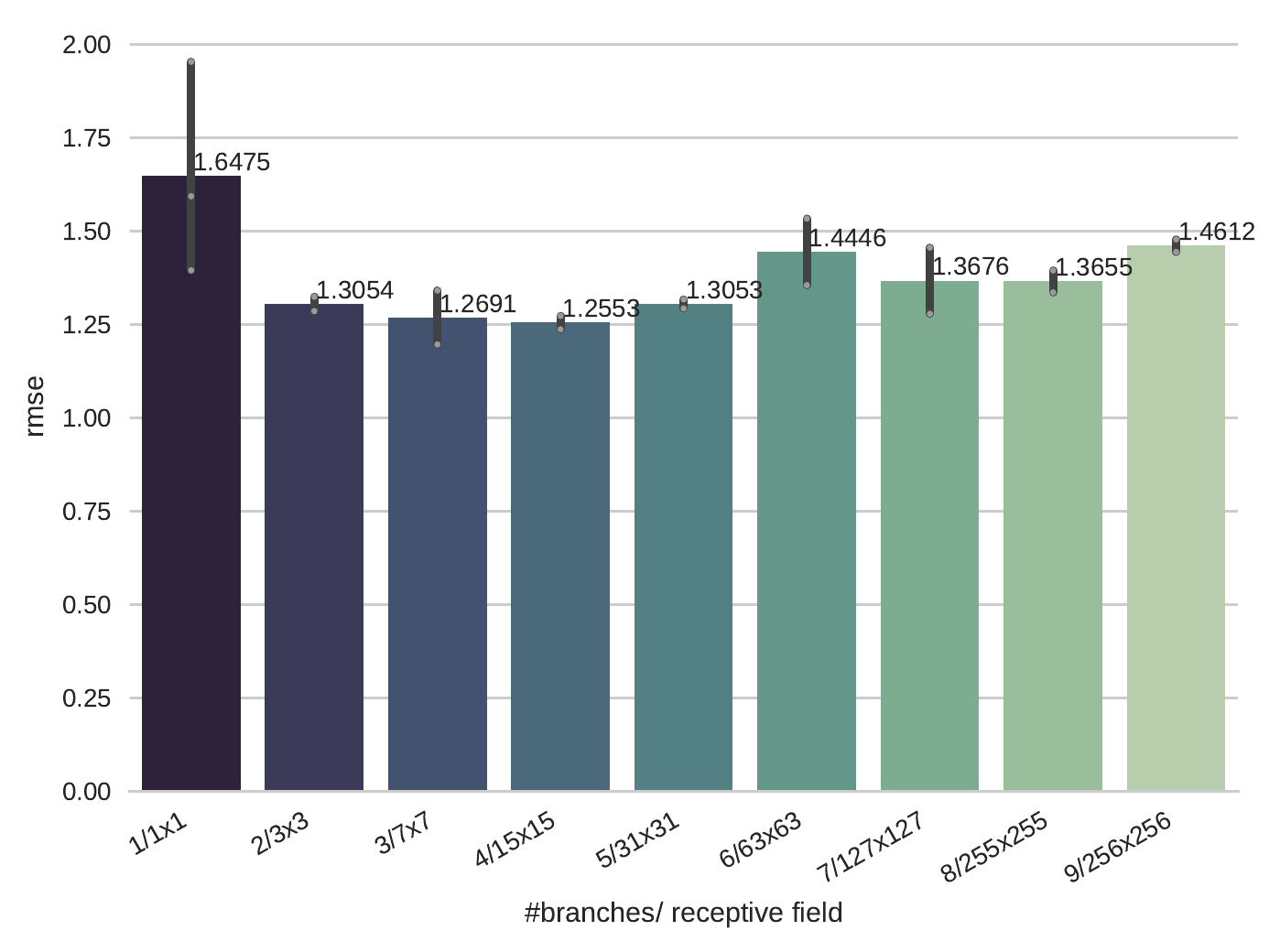}
        \includegraphics[width=0.495\linewidth]{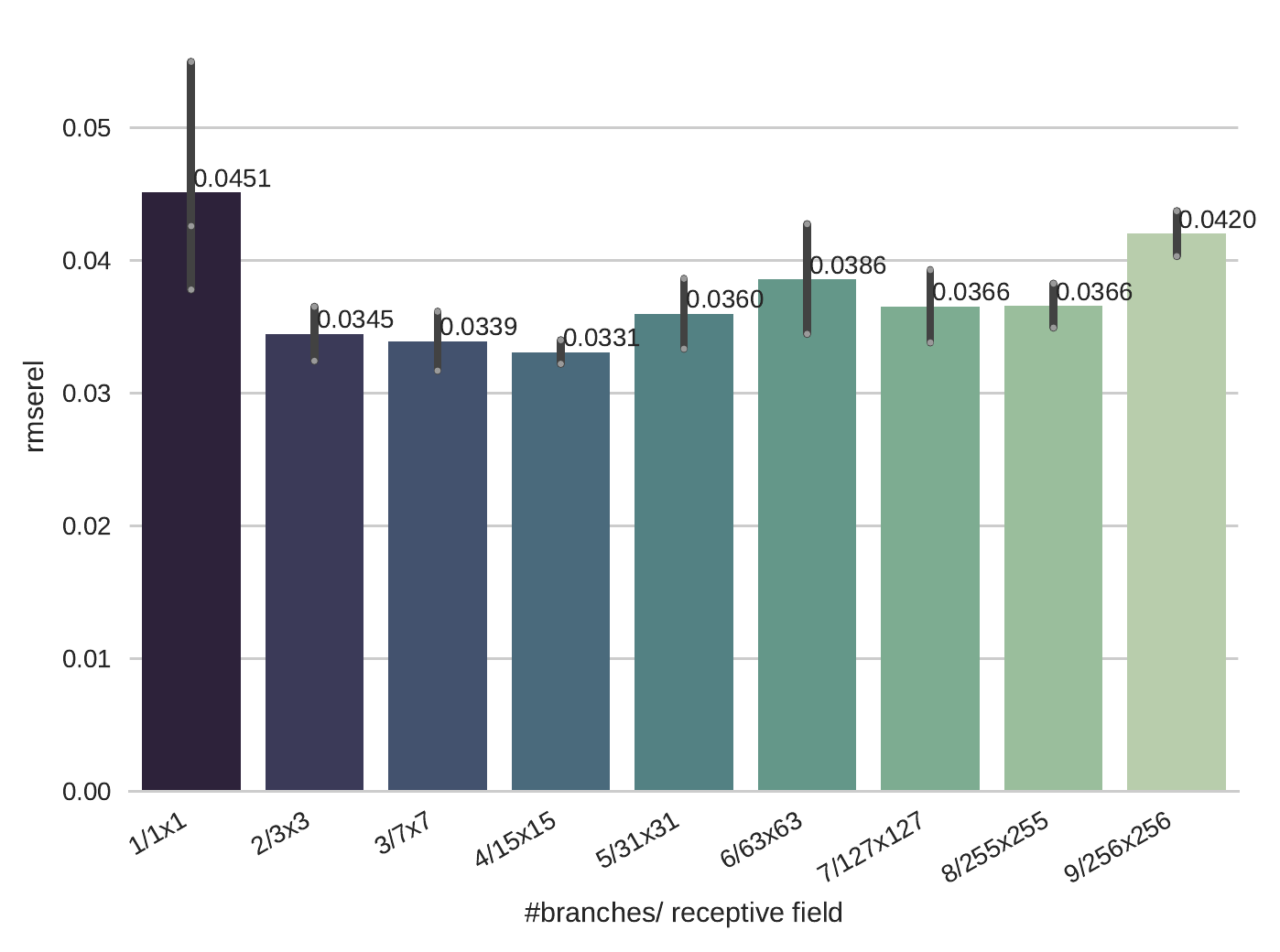}\\
        \includegraphics[width=0.495\linewidth]{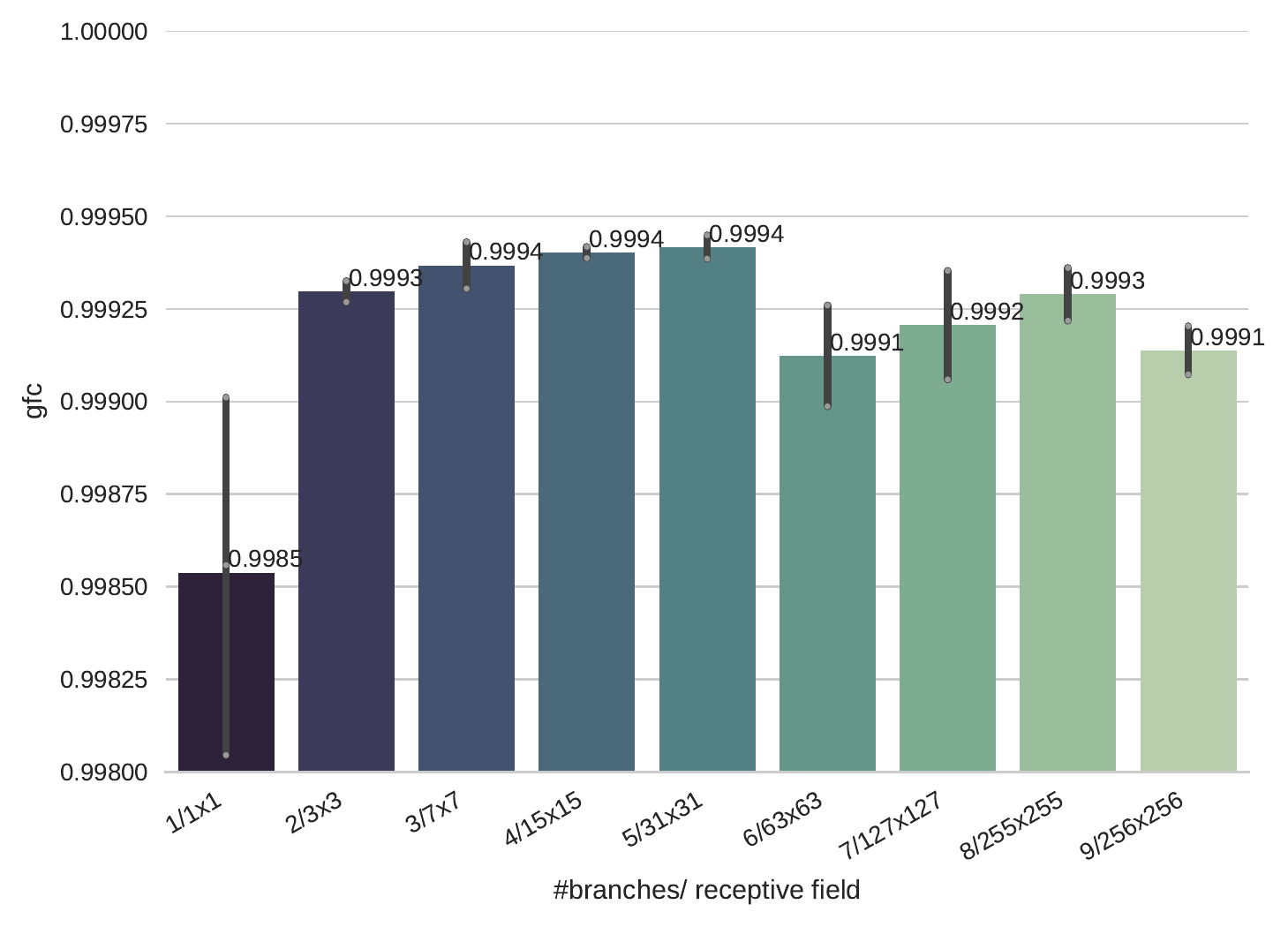}
        \includegraphics[width=0.495\linewidth]{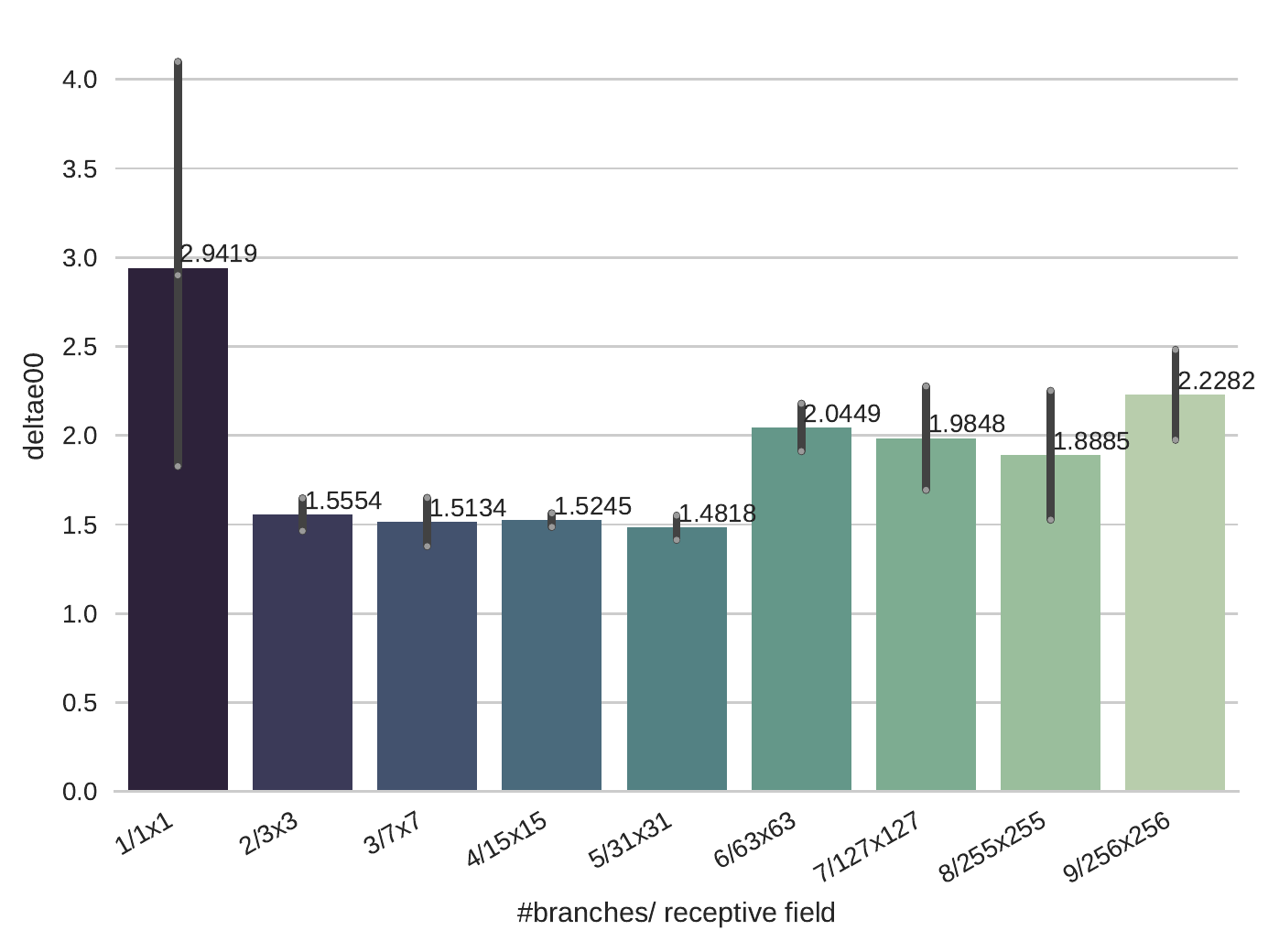}
	\end{center}
	\caption{Branch pruning experiment results. Top-left: RMSE. Top-right: RMSERel. Bottom-left: GFC. Bottom-right: $\Delta E_{00}$. Leftmost bar is the model with a single skip connection at $256\times256$ activation size level and $1\times1$ receptive field (RF). Each additional bar adds one skip connection at increasingly deeper levels of the U-Net. The rightmost bar is the full net, resulting from the addition of the main branch, and its RF (which would be $512\times512$ in an unconstrained scenario), is here limited by the $256\times256$ patch size. The addition of this last layer is justified by the notion of effective RF presented in \cite{luo_understanding_2016}, which may be significantly smaller than its theoretical counterpart.}
	\label{fig:pruning}
\end{figure*}
In an attempt to empirically validate our main hypothesis of contextual spatial information on a local neighborhood being relevant for the correct spectral reconstruction of any given central pixel, we conduct a branch pruning experiment. We depart from a minimal version of our net, in which both the main branch and all the skip connections have been removed, except for the one connecting the $256\times256$ input with the last pair of $1\times1$ convolutions (such model predicts each output pixel independently by design, without incorporating any spatial contribution), and keep adding skip connections at successively deeper levels (extending the receptive field of the model and thus increasing the spatial contribution at each step) until we end up with the full net, after the addition of the main $1\times1$ stream branch. Figure~\ref{fig:pruning} shows the results of running at least two train-test cycles on each of these nets, and testing over the $1280\times1280$ versions of the images in fold 1. All the four metrics show a closely correlated outcome, with a very significant average performance improvement ($-20.8\%$ RMSE, $-23.5\%$ RMSERel, $-47.1\% \Delta E_{00}$) when transitioning from the model with a single skip connection and a $1\times1$ receptive field to that with $2$ skip connections and a $3\times3$ receptive field. Further increases of the model's theoretical receptive field (by adding new branches) yield only marginally better results (models labeled as $3/ 7\times7$, $4/ 15\times15$) and, from there on, additional deeper skip connections produce increasing test error rates. We hypothesize that this is due to the influence of overfitting for experiments $5/ 31\times31$ and onwards. Given this, a straightforward way of improving the reported results could be that of increasing the regularization associated to the deepest branches by, e.g., increasing their dropout rate.

It is also noticeable the substantially higher variance that the results seem to suggest for the model with a $1\times1$ receptive field. This would mean that the addition of local spatial information does not only improve the overall prediction accuracy, but it does so in a more robust manner as well. 

\section{Conclusion}
\label{sec:conclusion}

We propose a convolutional neural network architecture that successfully learns an end-to-end mapping between pairs of input RGB images and their hyperspectral counterparts.
We adopt an adversarial framework-based generative model that shows itself effective in capturing the structure of the data manifold, and takes into account the spatial contextual information present in RGB images for the spectral reconstruction process. 
State of the art results in the ICVL dataset suggest that individual pixel-based approaches suffer from the fundamental limitation of not being able to effectively exploit the local context when applied to spectral image data in their attempt to build informative priors. 
The observed performance in terms of both reconstruction error and speed open the door to a full range of potential higher level applications in sectors of increasing demand for spectral footage at a lower cost.

\minisection{Acknowledgements}
This work has been partially funded by the Ministerio de Econom\'ia y Competitividad of Spain under grant number DPI2015-64571-R. We also acknowledge the Spanish project TIN2016-79717-R, and the CHIST-ERA project M2CR (PCIN-2015-251).  

{\small
\bibliographystyle{ieee}
\bibliography{adv_rgb2hs_better_bibtex_ascii_noforcekeys_bmvc2017}

\begin{thebibliography}{10}\itemsep=-1pt

\bibitem{agahian_reconstruction_2008}
F.~Agahian, S.~A. Amirshahi, and S.~H. Amirshahi.
\newblock Reconstruction of reflectance spectra using weighted principal
  component analysis.
\newblock {\em Color Research \& Application}, 33(5):360--371, Oct. 2008.

\bibitem{arad_sparse_2016}
B.~Arad and O.~Ben-Shahar.
\newblock Sparse {{Recovery}} of {{Hyperspectral Signal}} from {{Natural RGB
  Images}}.
\newblock In B.~Leibe, J.~Matas, N.~Sebe, and M.~Welling, editors, {\em
  Computer {{Vision}} -- {{ECCV}} 2016: 14th {{European Conference}},
  {{Amsterdam}}, {{The Netherlands}}, {{October}} 11--14, 2016,
  {{Proceedings}}, {{Part VII}}}, pages 19--34. {Springer International
  Publishing}, Cham, 2016.

\bibitem{ayala_use_2006}
F.~Ayala, J.~F. Ech{\'a}varri, P.~Renet, and A.~I. Negueruela.
\newblock Use of three tristimulus values from surface reflectance spectra to
  calculate the principal components for reconstructing these spectra by using
  only three eigenvectors.
\newblock {\em JOSA A}, 23(8):2020--2026, Aug. 2006.

\bibitem{bansal_pixelnet:_2017}
A.~Bansal, X.~Chen, B.~Russell, A.~Gupta, and D.~Ramanan.
\newblock {{PixelNet}}: {{Representation}} of the pixels, by the pixels, and
  for the pixels.
\newblock {\em arXiv:1702.06506 [cs]}, Feb. 2017.

\bibitem{cao_high_2011}
X.~Cao, X.~Tong, Q.~Dai, and S.~Lin.
\newblock High resolution multispectral video capture with a hybrid camera
  system.
\newblock In {\em {{CVPR}} 2011}, pages 297--304, June 2011.

\bibitem{chakrabarti2011statistics}
A.~Chakrabarti and T.~Zickler.
\newblock Statistics of {{Real}}-{{World Hyperspectral Images}}.
\newblock In {\em Proc.~{{IEEE Conf}}.~on {{Computer Vision}} and {{Pattern
  Recognition}} ({{CVPR}})}, pages 193--200, 2011.

\bibitem{Cheng_2015_ICCV}
Z.~Cheng, Q.~Yang, and B.~Sheng.
\newblock Deep {{Colorization}}.
\newblock In {\em The {{IEEE International Conference}} on {{Computer Vision}}
  ({{ICCV}})}, Dec. 2015.

\bibitem{cie_cie_2001}
{CIE}.
\newblock {{CIE}} 142-2001 {{Improvement}} to {{Industrial
  Colour}}-{{Difference Evaluation}}.
\newblock Technical Report CIE 142-2001, Commission Internationale de
  L'{\'e}clairage, Vienna, 2001.
\newblock ISBN 978 3 901906 08 4.

\bibitem{costa_towards_2017}
P.~Costa, A.~Galdran, M.~I. Meyer, M.~D. Abr{\`a}moff, M.~Niemeijer, A.~M.
  Mendon{\c c}a, and A.~Campilho.
\newblock Towards {{Adversarial Retinal Image Synthesis}}.
\newblock {\em arXiv:1701.08974 [cs, stat]}, Jan. 2017.

\bibitem{Dai_2016_CVPR}
J.~Dai, K.~He, and J.~Sun.
\newblock Instance-{{Aware Semantic Segmentation}} via {{Multi}}-{{Task Network
  Cascades}}.
\newblock In {\em The {{IEEE Conference}} on {{Computer Vision}} and {{Pattern
  Recognition}} ({{CVPR}})}, June 2016.

\bibitem{denton_deep_2015}
E.~L. Denton, S.~Chintala, a.~{szlam}, and R.~Fergus.
\newblock Deep {{Generative Image Models}} using a {{Laplacian Pyramid}} of
  {{Adversarial Networks}}.
\newblock In C.~Cortes, N.~D. Lawrence, D.~D. Lee, M.~Sugiyama, and R.~Garnett,
  editors, {\em Advances in {{Neural Information Processing Systems}} 28},
  pages 1486--1494. {Curran Associates, Inc.}, 2015.

\bibitem{eckhard_outdoor_2015}
J.~Eckhard, T.~Eckhard, E.~M. Valero, J.~L. Nieves, and E.~G. Contreras.
\newblock Outdoor scene reflectance measurements using a
  {{Bragg}}-grating-based hyperspectral imager.
\newblock {\em Applied Optics}, 54(13):D15--D24, May 2015.

\bibitem{foster_time-lapse_2016}
D.~H. Foster, K.~Amano, and S.~M. Nascimento.
\newblock Time-lapse ratios of cone excitations in natural scenes.
\newblock {\em Vision Research}, 120:45--60, Mar. 2016.

\bibitem{foster_frequency_2006}
D.~H. Foster, K.~Amano, S.~M.~C. Nascimento, and M.~J. Foster.
\newblock Frequency of metamerism in natural scenes.
\newblock {\em Journal of the Optical Society of America A}, 23(10):2359, Oct.
  2006.

\bibitem{galliani_learned_2017}
S.~Galliani, C.~Lanaras, D.~Marmanis, E.~Baltsavias, and K.~Schindler.
\newblock Learned {{Spectral Super}}-{{Resolution}}.
\newblock {\em arXiv:1703.09470 [cs]}, Mar. 2017.

\bibitem{goel_hypercam:_2015}
M.~Goel, E.~Whitmire, A.~Mariakakis, T.~S. Saponas, N.~Joshi, D.~Morris,
  B.~Guenter, M.~Gavriliu, G.~Borriello, and S.~N. Patel.
\newblock {{HyperCam}}: {{Hyperspectral Imaging}} for {{Ubiquitous Computing
  Applications}}.
\newblock In {\em Proceedings of the 2015 {{ACM International Joint
  Conference}} on {{Pervasive}} and {{Ubiquitous Computing}}}, UbiComp '15,
  pages 145--156, New York, NY, USA, 2015. {ACM}.

\bibitem{goodfellow_generative_2014}
I.~Goodfellow, J.~Pouget-Abadie, M.~Mirza, B.~Xu, D.~Warde-Farley, S.~Ozair,
  A.~Courville, and Y.~Bengio.
\newblock Generative {{Adversarial Nets}}.
\newblock In Z.~Ghahramani, M.~Welling, C.~Cortes, N.~D. Lawrence, and K.~Q.
  Weinberger, editors, {\em Advances in {{Neural Information Processing
  Systems}} 27}, pages 2672--2680. {Curran Associates, Inc.}, 2014.

\bibitem{heikkinen_evaluation_2008}
V.~Heikkinen, R.~Lenz, T.~Jetsu, J.~Parkkinen, M.~Hauta-Kasari, and
  T.~J{\"a}{\"a}skel{\"a}inen.
\newblock Evaluation and unification of some methods for estimating reflectance
  spectra from {{RGB}} images.
\newblock {\em JOSA A}, 25(10):2444--2458, Oct. 2008.

\bibitem{imai_comparative_2002}
F.~H. Imai, M.~R. Rosen, and R.~S. Berns.
\newblock Comparative study of metrics for spectral match quality.
\newblock In {\em Conference on {{Colour}} in {{Graphics}}, {{Imaging}}, and
  {{Vision}}}, volume 2002, pages 492--496. {Society for Imaging Science and
  Technology}, 2002.

\bibitem{ioffe_batch_2015}
S.~Ioffe and C.~Szegedy.
\newblock Batch {{Normalization}}: {{Accelerating Deep Network Training}} by
  {{Reducing Internal Covariate Shift}}.
\newblock In {\em Proceedings of the 32nd {{International Conference}} on
  {{Machine Learning}} ({{ICML}}-15)}, pages 448--456, 2015.

\bibitem{Isola_2017_CVPR}
P.~Isola, J.-Y. Zhu, T.~Zhou, and A.~A. Efros.
\newblock Image-{{To}}-{{Image Translation With Conditional Adversarial
  Networks}}.
\newblock In {\em The {{IEEE Conference}} on {{Computer Vision}} and {{Pattern
  Recognition}} ({{CVPR}})}, July 2017.

\bibitem{kawakami_high-resolution_2011}
R.~Kawakami, Y.~Matsushita, J.~Wright, M.~Ben-Ezra, Y.~W. Tai, and K.~Ikeuchi.
\newblock High-resolution hyperspectral imaging via matrix factorization.
\newblock In {\em {{CVPR}} 2011}, pages 2329--2336, June 2011.

\bibitem{keskar_large-batch_2017}
N.~S. Keskar, D.~Mudigere, J.~Nocedal, M.~Smelyanskiy, and P.~T.~P. Tang.
\newblock On {{Large}}-{{Batch Training}} for {{Deep Learning}}:
  {{Generalization Gap}} and {{Sharp Minima}}.
\newblock In {\em {{arXiv}}:1609.04836 [Cs, Math]}, Toulon, FR, Apr. 2017.

\bibitem{kingma_adam:_2015}
D.~P. Kingma and J.~Ba.
\newblock Adam: {{A Method}} for {{Stochastic Optimization}}.
\newblock In {\em Proceedings of the 3rd {{International Conference}} on
  {{Learning Representations}} ({{ICLR}})}, San Diego, USA, 2015.

\bibitem{kingma_auto-encoding_2013}
D.~P. Kingma and M.~Welling.
\newblock Auto-{{Encoding Variational Bayes}}.
\newblock In {\em Proceedings of the 2nd {{International Conference}} on
  {{Learning Representations}} ({{ICLR}})}, 2013.

\bibitem{larsen_autoencoding_2016}
A.~B.~L. Larsen, S.~K. S{\o}nderby, H.~Larochelle, and O.~Winther.
\newblock Autoencoding beyond pixels using a learned similarity metric.
\newblock In M.~F. Balcan and K.~Q. Weinberger, editors, {\em Proceedings of
  {{The}} 33rd {{International Conference}} on {{Machine Learning}}}, volume~48
  of {\em Proceedings of Machine Learning Research}, pages 1558--1566, New
  York, New York, USA, 20--22 Jun 2016. {PMLR}.

\bibitem{Li2016}
C.~Li and M.~Wand.
\newblock Precomputed {{Real}}-{{Time Texture Synthesis}} with {{Markovian
  Generative Adversarial Networks}}.
\newblock In B.~Leibe, J.~Matas, N.~Sebe, and M.~Welling, editors, {\em
  Computer {{Vision}} -- {{ECCV}} 2016: 14th {{European Conference}},
  {{Amsterdam}}, {{The Netherlands}}, {{October}} 11-14, 2016, {{Proceedings}},
  {{Part III}}}, pages 702--716. {Springer International Publishing}, Cham,
  2016.

\bibitem{lopez-alvarez_using_2008}
M.~A. L{\'o}pez-{\'A}lvarez, J.~Hern{\'a}ndez-Andr{\'e}s, J.~Romero, F.~J.
  Olmo, A.~Cazorla, and L.~Alados-Arboledas.
\newblock Using a trichromatic {{CCD}} camera for spectral skylight estimation.
\newblock {\em Applied Optics}, 47(34):H31--H38, Dec. 2008.

\bibitem{luo_understanding_2016}
W.~Luo, Y.~Li, R.~Urtasun, and R.~Zemel.
\newblock Understanding the {Effective} {Receptive} {Field} in {Deep}
  {Convolutional} {Neural} {Networks}.
\newblock In D.~D. Lee, U.~V. Luxburg, I.~Guyon, and R.~Garnett, editors, {\em
  Advances {In} {Neural} {Information} {Processing} {Systems} 29}, pages
  4898--4906. Curran Associates, Inc., 2016.

\bibitem{mathieu2015deep}
M.~Mathieu, C.~Couprie, and Y.~LeCun.
\newblock Deep multi-scale video prediction beyond mean square error.
\newblock In {\em International {{Conference}} on {{Learning Representations}}
  ({{ICLR}} 2016)}. {arXiv preprint arXiv:1511.05440}, 2015.

\bibitem{mirza2014conditional}
M.~Mirza and S.~Osindero.
\newblock Conditional generative adversarial nets.
\newblock In {\em {{arXiv}} Preprint {{arXiv}}:1411.1784}, 2014.

\bibitem{moan_image-difference_2014}
S.~L. Moan and P.~Urban.
\newblock Image-{{Difference Prediction}}: {{From Color}} to {{Spectral}}.
\newblock {\em IEEE Transactions on Image Processing}, 23(5):2058--2068, May
  2014.

\bibitem{nguyen_training-based_2014}
R.~M.~H. Nguyen, D.~K. Prasad, and M.~S. Brown.
\newblock Training-{{Based Spectral Reconstruction}} from a {{Single RGB
  Image}}.
\newblock In D.~Fleet, T.~Pajdla, B.~Schiele, and T.~Tuytelaars, editors, {\em
  Computer {{Vision}} {{ECCV}} 2014}, number 8695 in Lecture Notes in Computer
  Science, pages 186--201. {Springer International Publishing}, Sept. 2014.

\bibitem{park_multispectral_2007}
J.~I. Park, M.~H. Lee, M.~D. Grossberg, and S.~K. Nayar.
\newblock Multispectral {{Imaging Using Multiplexed Illumination}}.
\newblock In {\em 2007 {{IEEE}} 11th {{International Conference}} on {{Computer
  Vision}}}, pages 1--8, Oct. 2007.

\bibitem{parmar_spatio-spectral_2008}
M.~Parmar, S.~Lansel, and B.~A. Wandell.
\newblock Spatio-spectral reconstruction of the multispectral datacube using
  sparse recovery.
\newblock In {\em 2008 15th {{IEEE International Conference}} on {{Image
  Processing}}}, pages 473--476, Oct. 2008.

\bibitem{Pathak_2016_CVPR}
D.~Pathak, P.~Krahenbuhl, J.~Donahue, T.~Darrell, and A.~A. Efros.
\newblock Context {{Encoders}}: {{Feature Learning}} by {{Inpainting}}.
\newblock In {\em The {{IEEE Conference}} on {{Computer Vision}} and {{Pattern
  Recognition}} ({{CVPR}})}, June 2016.

\bibitem{radford_unsupervised_2016}
A.~Radford, L.~Metz, and S.~Chintala.
\newblock Unsupervised {{Representation Learning}} with {{Deep Convolutional
  Generative Adversarial Networks}}.
\newblock In {\em Proceedings of the 4th {{International Conference}} on
  {{Learning Representations}} ({{ICLR}})}, 2016.

\bibitem{reed_generative_2016}
S.~Reed, Z.~Akata, X.~Yan, L.~Logeswaran, H.~Lee, and B.~Schiele.
\newblock Generative {{Adversarial Text}} to {{Image Synthesis}}.
\newblock In {\em International {{Conference}} on {{Machine Learning}}
  ({{ICML}})}, 2016.

\bibitem{reed_learning_2016}
S.~E. Reed, Z.~Akata, S.~Mohan, S.~Tenka, B.~Schiele, and H.~Lee.
\newblock Learning {{What}} and {{Where}} to {{Draw}}.
\newblock In D.~D. Lee, M.~Sugiyama, U.~V. Luxburg, I.~Guyon, and R.~Garnett,
  editors, {\em Advances in {{Neural Information Processing Systems}} 29},
  pages 217--225. {Curran Associates, Inc.}, 2016.

\bibitem{RoblesKelly2015SingleIS}
A.~Robles-Kelly.
\newblock Single {{Image Spectral Reconstruction}} for {{Multimedia
  Applications}}.
\newblock In {\em {{ACM Multimedia}}}, 2015.

\bibitem{romero1997linear}
J.~Romero, A.~Garc{\i}a-Beltr{\'a}n, and J.~Hern{\'a}ndez-Andr{\'e}s.
\newblock Linear bases for representation of natural and artificial
  illuminants.
\newblock {\em JOSA A}, 14(5):1007--1014, 1997.

\bibitem{ronneberger_u-net:_2015}
O.~Ronneberger, P.~Fischer, and T.~Brox.
\newblock U-{{Net}}: {{Convolutional Networks}} for {{Biomedical Image
  Segmentation}}.
\newblock In {\em Medical {{Image Computing}} and {{Computer}}-{{Assisted
  Intervention}} \textendash{} {{MICCAI}} 2015}, pages 234--241. {Springer,
  Cham}, Oct. 2015.

\bibitem{salimans_improved_2016}
T.~Salimans, I.~Goodfellow, W.~Zaremba, V.~Cheung, A.~Radford, X.~Chen, and
  X.~Chen.
\newblock Improved {{Techniques}} for {{Training GANs}}.
\newblock In D.~D. Lee, M.~Sugiyama, U.~V. Luxburg, I.~Guyon, and R.~Garnett,
  editors, {\em Advances in {{Neural Information Processing Systems}} 29},
  pages 2226--2234. {Curran Associates, Inc.}, 2016.

\bibitem{shafer_using_1985}
S.~A. Shafer.
\newblock Using color to separate reflection components.
\newblock {\em Color Research \& Application}, 10(4):210--218, Dec. 1985.

\bibitem{shelhamer_fully_2016}
E.~Shelhamer, J.~Long, and T.~Darrell.
\newblock Fully {{Convolutional Networks}} for {{Semantic Segmentation}}.
\newblock {\em IEEE Transactions on Pattern Analysis and Machine Intelligence},
  PP(99):1--1, 2016.

\bibitem{Shrivastava_2017_CVPR}
A.~Shrivastava, T.~Pfister, O.~Tuzel, J.~Susskind, W.~Wang, and R.~Webb.
\newblock Learning {{From Simulated}} and {{Unsupervised Images Through
  Adversarial Training}}.
\newblock In {\em The {{IEEE Conference}} on {{Computer Vision}} and {{Pattern
  Recognition}} ({{CVPR}})}, July 2017.

\bibitem{theis_note_2016}
L.~Theis, A.~van~den Oord, and M.~Bethge.
\newblock A note on the evaluation of generative models.
\newblock In {\em Proceedings of the 4th {{International Conference}} on
  {{Learning Representations}} ({{ICLR}})}, 2016.

\bibitem{varma_classifying_2002}
M.~Varma and A.~Zisserman.
\newblock Classifying {{Images}} of {{Materials}}: {{Achieving Viewpoint}} and
  {{Illumination Independence}}.
\newblock In {\em Computer {{Vision}} \textemdash{} {{ECCV}} 2002}, Lecture
  Notes in Computer Science, pages 255--271. {Springer, Berlin, Heidelberg},
  May 2002.

\bibitem{wang_generative_2016}
X.~Wang and A.~Gupta.
\newblock Generative {{Image Modeling Using Style}} and {{Structure Adversarial
  Networks}}.
\newblock In B.~Leibe, J.~Matas, N.~Sebe, and M.~Welling, editors, {\em
  Computer {{Vision}} -- {{ECCV}} 2016: 14th {{European Conference}},
  {{Amsterdam}}, {{The Netherlands}}, {{October}} 11--14, 2016,
  {{Proceedings}}, {{Part IV}}}, pages 318--335. {Springer International
  Publishing}, Cham, 2016.

\bibitem{yasuma_generalized_2010}
F.~Yasuma, T.~Mitsunaga, D.~Iso, and S.~K. Nayar.
\newblock Generalized {{Assorted Pixel Camera}}: {{Postcapture Control}} of
  {{Resolution}}, {{Dynamic Range}}, and {{Spectrum}}.
\newblock {\em IEEE Transactions on Image Processing}, 19(9):2241--2253, Sept.
  2010.

\bibitem{zhang_colorful_2016}
R.~Zhang, P.~Isola, and A.~A. Efros.
\newblock Colorful {{Image Colorization}}.
\newblock In B.~Leibe, J.~Matas, N.~Sebe, and M.~Welling, editors, {\em
  Computer {{Vision}} -- {{ECCV}} 2016: 14th {{European Conference}},
  {{Amsterdam}}, {{The Netherlands}}, {{October}} 11-14, 2016, {{Proceedings}},
  {{Part III}}}, pages 649--666. {Springer International Publishing}, Cham,
  2016.

\bibitem{zhao_image-based_2007}
Y.~Zhao and R.~S. Berns.
\newblock Image-based spectral reflectance reconstruction using the matrix
  {{R}} method.
\newblock {\em Color Research \& Application}, 32(5):343--351, Oct. 2007.

\bibitem{zhu_generative_2016}
J.-Y. Zhu, P.~Kr{\"a}henb{\"u}hl, E.~Shechtman, and A.~A. Efros.
\newblock Generative {{Visual Manipulation}} on the {{Natural Image Manifold}}.
\newblock In B.~Leibe, J.~Matas, N.~Sebe, and M.~Welling, editors, {\em
  Computer {{Vision}} -- {{ECCV}} 2016: 14th {{European Conference}},
  {{Amsterdam}}, {{The Netherlands}}, {{October}} 11-14, 2016, {{Proceedings}},
  {{Part V}}}, pages 597--613. {Springer International Publishing}, Cham, 2016.

\end{thebibliography}
}

\clearpage

\section{Appendix}
\subsection{Train-test splits}
The following lists contain the file identifiers belonging to each of the sets used in our experiments.

\begin{itemize}
\item \textbf{Fold $0$ train set $=$ Fold $1$ test set:}
\begin{lstlisting}
4cam_0411-1648.mat
bguCAMP_0514-1659.mat
bguCAMP_0514-1711.mat
bguCAMP_0514-1712.mat
BGU_0403-1419-1.mat
bgu_0403-1459.mat
bgu_0403-1525.mat
BGU_0522-1201.mat
BGU_0522-1211.mat
BGU_0522-1217.mat
bulb_0822-0903.mat
CC_40D_2_1103-0917.mat
eve_0331-1549.mat
eve_0331-1601.mat
eve_0331-1602.mat
eve_0331-1618.mat
eve_0331-1633.mat
eve_0331-1646.mat
eve_0331-1647.mat
eve_0331-1656.mat
eve_0331-1657.mat
Flower_0325-1336.mat
gavyam_0823-0930.mat
Maz0326-1038.mat
maz_0326-1048.mat
mor_0328-1209-2.mat
nachal_0823-1040.mat
nachal_0823-1047.mat
nachal_0823-1110.mat
nachal_0823-1117.mat
nachal_0823-1127.mat
nachal_0823-1147.mat
nachal_0823-1149.mat
nachal_0823-1152.mat
nachal_0823-1210-4.mat
nachal_0823-1213.mat
nachal_0823-1222.mat
negev_0823-1003.mat
objects_0924-1557.mat
objects_0924-1558.mat
objects_0924-1601.mat
objects_0924-1602.mat
objects_0924-1605.mat
objects_0924-1612.mat
objects_0924-1617.mat
objects_0924-1619.mat
objects_0924-1620.mat
objects_0924-1625.mat
objects_0924-1627.mat
objects_0924-1631.mat
objects_0924-1632.mat
objects_0924-1633.mat
objects_0924-1636.mat
objects_0924-1652.mat
omer_0331-1118.mat
omer_0331-1119.mat
omer_0331-1130.mat
omer_0331-1131.mat
omer_0331-1135.mat
omer_0331-1150.mat
peppers_0503-1311.mat
peppers_0503-1315.mat
peppers_0503-1330.mat
pepper_0503-1228.mat
pepper_0503-1229.mat
pepper_0503-1236.mat
plt_0411-1116.mat
prk_0328-0945.mat
prk_0328-1031.mat
Ramot0325-1364.mat
rmt_0328-1241-1.mat
rmt_0328-1249-1.mat
rsh_0406-1343.mat
rsh_0406-1413.mat
sat_0406-1107.mat
sat_0406-1129.mat
strt_0331-1027.mat
tree_0822-0853.mat
gavyam_0823-0944.mat
gavyam_0823-0945.mat
hill_0325-1228.mat
hill_0325-1235.mat
IDS_COLORCHECK_1020-1223.mat
Labtest_0910-1502.mat
Labtest_0910-1504.mat
Labtest_0910-1509.mat
Labtest_0910-1510.mat
Labtest_0910-1511.mat
lehavim_0910-1600.mat
Lehavim_0910-1622.mat
Lehavim_0910-1626.mat
Lehavim_0910-1633.mat
Lehavim_0910-1635.mat
Lehavim_0910-1636.mat
Lehavim_0910-1716.mat
Lehavim_0910-1717.mat
Lehavim_0910-1718.mat
gavyam_0823-0933.mat
Master2900k.mat
objects_0924-1641.mat
\end{lstlisting}
\newpage
\item \textbf{Fold $1$ train set $=$ Fold $0$ test set:}
\begin{lstlisting}
4cam_0411-1640-1.mat
bguCAMP_0514-1718.mat
bguCAMP_0514-1723.mat
bguCAMP_0514-1724.mat
bgu_0403-1439.mat
bgu_0403-1444.mat
bgu_0403-1511.mat
bgu_0403-1523.mat
BGU_0522-1113-1.mat
BGU_0522-1127.mat
BGU_0522-1136.mat
BGU_0522-1203.mat
BGU_0522-1216.mat
bulb_0822-0909.mat
eve_0331-1551.mat
eve_0331-1606.mat
eve_0331-1632.mat
eve_0331-1702.mat
eve_0331-1705.mat
gavyam_0823-0950-1.mat
grf_0328-0949.mat
hill_0325-1219.mat
hill_0325-1242.mat
IDS_COLORCHECK_1020-1215-1.mat
Labtest_0910-1506.mat
Labtest_0910-1513.mat
lehavim_0910-1602.mat
lehavim_0910-1605.mat
lehavim_0910-1607.mat
lehavim_0910-1610.mat
objects_0924-1610.mat
objects_0924-1611.mat
objects_0924-1622.mat
objects_0924-1623.mat
objects_0924-1624.mat
objects_0924-1628.mat
objects_0924-1629.mat
objects_0924-1634.mat
objects_0924-1637.mat
objects_0924-1638.mat
objects_0924-1639.mat
objects_0924-1645.mat
objects_0924-1648.mat
objects_0924-1650.mat
omer_0331-1055.mat
omer_0331-1102.mat
omer_0331-1104.mat
omer_0331-1159.mat
peppers_0503-1308.mat
peppers_0503-1332.mat
plt_0411-1037.mat
plt_0411-1046.mat
Lehavim_0910-1629.mat
Lehavim_0910-1630.mat
Lehavim_0910-1640.mat
Lehavim_0910-1708.mat
Lehavim_0910-1725.mat
lst_0408-0950.mat
lst_0408-1004.mat
lst_0408-1012.mat
Master20150112_f2_colorchecker.mat
Master5000K.mat
Master5000K_2900K.mat
nachal_0823-1038.mat
nachal_0823-1118.mat
nachal_0823-1121.mat
nachal_0823-1132.mat
nachal_0823-1144.mat
nachal_0823-1145.mat
nachal_0823-1214.mat
nachal_0823-1217.mat
nachal_0823-1220.mat
nachal_0823-1223.mat
negev_0823-1005.mat
objects_0924-1550.mat
objects_0924-1556.mat
objects_0924-1600.mat
plt_0411-1200-1.mat
plt_0411-1207.mat
plt_0411-1210.mat
plt_0411-1211.mat
plt_0411-1232-1.mat
prk_0328-1025.mat
prk_0328-1034.mat
prk_0328-1037.mat
prk_0328-1045.mat
ramot_0325-1322.mat
rsh2_0406-1505.mat
rsh_0406-1356.mat
rsh_0406-1427.mat
rsh_0406-1441-1.mat
rsh_0406-1443.mat
sami_0331-1019.mat
sat_0406-1130.mat
sat_0406-1157-1.mat
selfie_0822-0906.mat
ulm_0328-1118.mat
Lehavim_0910-1627.mat
objects_0924-1607.mat
plt_0411-1155.mat
\end{lstlisting}
\end{itemize}

\end{document}


\maketitle


\appendix
\section{Network arquitectures of Generator and Discriminator}

\begin{figure*}
	\begin{center}
        \fbox{\includegraphics[width=0.2\linewidth]{images/eg1_largeprint.png}}
	\end{center}
	\caption{Schematic of \emph{U-Net} architecture.\textbf{TODO}}
	\label{fig:unet}
\end{figure*}

\section{evaluation metrics}

\begin{equation} \label{eq:gfc_def}
GFC = \dfrac{1}{N}\sum_x\frac{|\displaystyle\sum_{\lambda} S(\lambda, x)\hat{S}(\lambda,x)|}{\sqrt{\displaystyle\sum_{\lambda}\big[ S(\lambda, x)\big]^2} \sqrt{\displaystyle\sum_{\lambda}\big[\hat{S}(\lambda, x)\big]^2}}
\end{equation}

\section{dataset preparation}
\begin{equation} \label{eq:preproc_smin}
S''(\lambda, x) = S'(\lambda, x) - \min_{train\_set}S'(\lambda, x)
\end{equation}
\begin{equation} \label{eq:preproc_smax}
S(\lambda, x) = \frac{S''(\lambda, x)}{\displaystyle \max_{train\_set}S''(\lambda, x)}
\end{equation}

	\subsubsection{Experiment from issue \#94}
    Experiment: train on the whole db except for the specific sets [park(9) / indoor(2) / urban(36) / Rural(4) / Plant(4)] and evaluate on these.
	\subsubsection{Experiment from issue \#95}: 
Experiment: reproduce the cross domain results from Table 1: park(9) <->rural(4) with very few images

TIME PERFORMANCE IT IS FAST
{\bf Include review discussion on individual spectra and image matching metrics}

MESSAGES:
non uniform errors are very harmful because they alter the color response.
\bibliography{adv_rgb2hs_supl}